\documentclass{article} %
\usepackage{iclr2022_conference,times}

\usepackage{amsmath,amsfonts,bm}

\def\eqref#1{equation~\ref{#1}}

\def\1{\bm{1}}

\DeclareMathAlphabet{\mathsfit}{\encodingdefault}{\sfdefault}{m}{sl}
\SetMathAlphabet{\mathsfit}{bold}{\encodingdefault}{\sfdefault}{bx}{n}

\usepackage[T1]{fontenc}
\usepackage{mathtools}
\usepackage{amsmath}
\usepackage{amssymb}
\usepackage{enumerate}
\usepackage{graphicx,subcaption}
\usepackage{xspace}
\usepackage[font=small]{caption} 

\usepackage{booktabs}
\usepackage{multirow}

\usepackage{hyperref}
\usepackage{url}
\newcommand*{\yoruba}{Yor\`ub\'a\xspace}

\title{Task-Adaptive Pre-Training for Boosting Learning With Noisy Labels: A Study on Text Classification for African Languages}

\author{Dawei Zhu, Michael A. Hedderich, Fangzhou Zhai, David Ifeoluwa Adelani, Dietrich Klakow \\
Saarland University, Saarland Informatics Campus, Germany\\
\texttt{\{dzhu,mhedderich,didelani,dietrich.klakow\}@lsv.uni-saarland.de} \\
\texttt{fzhai@coli.uni-saarland.de}
}

\iclrfinalcopy %
\begin{document}

\maketitle

\vspace{-0.5cm}
For high-resource languages like English, text classification is a well-studied task. The performance of modern NLP models easily achieves an accuracy of more than 90\% in many standard datasets for text classification in English \citep{xie2019unsupervised, Yang2019, Zaheer2020}. However, text classification in low-resource languages is still challenging due to the lack of annotated data. Although methods like weak supervision and crowdsourcing can help ease the annotation bottleneck, the annotations obtained by these methods contain label noise. Models trained with label noise may not generalize well. To this end, a variety of noise-handling techniques have been proposed to alleviate the negative impact caused by the errors in the annotations (for extensive surveys see \citep{hedderich-etal-2021-survey, DBLP:journals/kbs/AlganU21}). In this work, we experiment with a group of standard noisy-handling methods on text classification tasks with noisy labels. We study both simulated noise and realistic noise induced by weak supervision. Moreover, we find task-adaptive pre-training techniques \citep{DBLP:conf/acl/GururanganMSLBD20} are beneficial for learning with noisy labels. 

\paragraph{Problem Settings} We consider a $k$-class text classification problem with noisy labels. In particular, instead of accessing a training set $D=\{(x_i, y_i)_{i=1}^n\}$ sampled from the clean data generation distribution, we work with a noisy version of $D$, denoted by $\hat{D}=\{(x_i, \hat{y}_i)_{i=1}^n\}$. The goal is to train a model on $\hat{D}$ that generalizes well on the clean distribution.
\paragraph{Noise Simulation}
Research in this field often constructs a noisy dataset by injecting noise into a clean dataset \citep{DBLP:conf/nips/HanYYNXHTS18, DBLP:conf/naacl/JindalPLN19, DBLP:journals/corr/abs-2105-00828}. This simulates annotation scenarios such as crowdsourcing, where some annotators answer randomly or overlook some entries in multiple-choice questions. It also allows controlling the noise level and type. The noise in such a simulation is feature-independent. That is, it assumes $p(\hat{y}|y=i,x) = p(\hat{y}|y=i)$.
\paragraph{Weak Supervision Noise}
In weak supervision, the labels are annotated in an automatic manner. For example, \citet{DBLP:conf/emnlp/HedderichAZAMK20} assign labels to a text document by leveraging word lists and a set of simple rules. Unlike the simulated noise, the noise caused by weak supervision is feature-dependent, meaning that the independence assumption does not hold.

\paragraph{Task-Adaptive Pre-Training (TAPT)}
The BERT model is pre-trained on a large unlabeled, general-domain corpus. Task-adaptive pre-training further pre-trains BERT on text from the downstream task. In particular, given a task-domain corpus, one continues to train BERT with the Masked Language Model (MLM) task and the Next Sentence Prediction (NSP) task. Previous studies have shown that task-adaptive pre-training is beneficial for downstream NLP tasks with clean training data \citep{DBLP:conf/cncl/SunQXH19,DBLP:conf/acl/GururanganMSLBD20}. However, it remains unclear whether the noisy settings can inherit this benefit.

\paragraph{Data}
We simulate the noise on two established text classification dataset: AG-News \citep{DBLP:conf/nips/ZhangZL15} and IMDB \citep{DBLP:conf/acl/MaasDPHNP11}. We study two simulated noise types, uniform \citep{DBLP:conf/nips/RooyenMW15} and single-flip \citep{DBLP:journals/corr/ReedLASER14} noise. We also evaluate two datasets with weak supervision noise in two low-resource African languages: Hausa and \yoruba \citep{DBLP:conf/emnlp/HedderichAZAMK20}. The weak labels for these two datasets are generated by simple rules. For example, to identify texts for the class ``Africa'', a labeling rule based on a list of African countries and their capitals is used.

\vspace{-0.3cm}
\paragraph{Baselines}
We compare learning without noise-handling with four popular noise-handling methods. 1) \textbf{Co-Teaching}: \citet{DBLP:conf/nips/HanYYNXHTS18} trains two networks to pick cleaner training subsets for each other. 2) \textbf{Noise Matrix}: A noise transition matrix is appended after BERT's prediction to transform the clean label distribution to the noisy one. A variety of methods exists for estimating the noise matrix, i.e. \citet{sukhbaatar2015training, DBLP:conf/icassp/BekkerG16, DBLP:conf/cvpr/PatriniRMNQ17, DBLP:conf/nips/HendrycksMWG18, DBLP:conf/nips/YaoL0GD0S20}. To exclude the effects of estimation errors in the evaluation, we use the ground truth transition matrix as it is the best possible estimation. This matrix is fixed after initialization. 3): \textbf{Noise Matrix with Regularization}:  Similar to \textit{Noise Matrix}, \citet{DBLP:conf/naacl/JindalPLN19} appends a noise matrix after BERT's output. During training, the matrix is learned with an $l2$ regularization and is not necessarily normalized to be a probability matrix. This method is the current state-of-the-art method for text classification with simulated noise 4) \textbf{Label Smoothing}: label smoothing \cite{Szegedy2016RethinkingTI} is a commonly used method to improve model's generalization and calibration. It mixes the one-hot label with a uniform vector, preventing the model from getting overconfident on the samples. \citet{lukasik2020does} further shows that it improves noise robustness. Finally, for the sake of comparison, we also include a model, denoted by \textbf{No Validation}, which neither applies noise-handling nor early-stopping.

\begin{table}[t]
    \scriptsize
    \begin{subtable}[t]{0.5\textwidth}
        \centering
        \begin{tabular}{ccc}
            \toprule
            Methods & Original & +TAPT\\ \midrule
            Without Noise-handling  & 85.49 & 2.13$\uparrow$ \\
            Co-Teaching & 84.74 & 0.90$\uparrow$ \\ 
            Noise Matrix & 83.90 & 2.80$\uparrow$ \\
            Noise Matrix with Regularization & 84.77 & 2.77$\uparrow$ \\
            Label Smoothing & 86.64 & 0.59$\uparrow$ \\
            \bottomrule
        \end{tabular}
        \caption{AG-News, 70\% uniform noise}
    \end{subtable}
    \begin{subtable}[t]{0.5\textwidth}
        \centering
        \begin{tabular}{ccc}
            \toprule
            Methods & Original & +TAPT\\ \midrule
            Without Noise-handling  & 80.12 & 4.54$\uparrow$ \\
            Co-Teaching & 83.77 & 6.14$\uparrow$ \\ 
            Noise Matrix & 78.82 & 2.80$\uparrow$ \\
            Noise Matrix with Regularization & 80.17 & 4.15$\uparrow$ \\
            Label Smoothing & 80.61 & 0.10$\uparrow$ \\
            \bottomrule
        \end{tabular}
        \caption{IMDB, 45\% uniform noise}
    \end{subtable}
    \newline
    \vspace*{0.5cm}
    \newline
        \begin{subtable}[t]{0.5\textwidth}
        \centering
        \begin{tabular}{ccc}
            \toprule
            Methods & Original & +TAPT\\ \midrule
            Without Noise-handling  & 64.72 & 4.86$\uparrow$ \\
            Co-Teaching & 61.37 & 1.98$\uparrow$ \\ 
            Noise Matrix & 65.96 & 3.12$\uparrow$ \\
            Noise Matrix with Regularization & 61.32 & 2.14$\uparrow$ \\
            Label Smoothing & 65.44 & 0.59$\uparrow$ \\
            \bottomrule
        \end{tabular}
        \caption{\yoruba, weak supervision noise}
    \end{subtable}
    \begin{subtable}[t]{0.5\textwidth}
        \centering
        \begin{tabular}{ccc}
            \toprule
            Methods & Original & +TAPT\\ \midrule
            Without Noise-handling  & 46.97 & 0.85$\uparrow$ \\
            Co-Teaching & 31.65 & 4.51$\uparrow$ \\ 
            Noise Matrix & 46.58 & 0.79$\uparrow$ \\
            Noise Matrix with Regularization & 35.36 & 5.71$\uparrow$ \\
            Label Smoothing & 46.44 & 0.51$\uparrow$ \\
            \bottomrule
        \end{tabular}
        \caption{Hausa, weak supervision noise}
    \end{subtable}

    \caption{Performance of the original models and that after applying TAPT to the BERT backbone. TAPT consistently benifits all models in all noise settings. Performance measured in Accuracy (\%). The mean performance over five trials is reported.}
    \label{table:TPAT_difference}
\end{table}

\vspace{-0.3cm}
\paragraph{Results} We evaluate our baselines on four datasets with different noise levels and noise types. We find that BERT is robust to simulated noise, especially under low to mild noise levels. For example, the performance of BERT without noise-handling drops less than 5\% under 60\% uniform noise in AG-News. In these cases, noise-handling methods rarely outperform the baseline without noise-handling. When the noise levels go higher, noise-handling starts benefiting learning. However, the results are mixed. For example, Co-Teaching outperforms the baseline without noise-handling on IMDB with 45\% single-flip noise, yet it underperforms in AG-News with 70\% uniform noise. Please refer to Appendix \ref{appendix:detailed_performance} for detailed performance numbers in each noise setting.

Task-adaptive pre-training offers an implicit way to assist BERT in all noise settings we investigate. Table \ref{table:TPAT_difference} presents the performance difference before and after adding TAPT. Compared to the mixed results we see from applying noise-handling, TAPT helps consistently. We also notice that the performance difference between different methods is reduced. This could be particularly beneficial if computational resources are limited and trying out all noise-handling methods is not possible. In this case, just using TAPT without a complex noise-handling architecture should already offer a strong baseline.

In summary, we observe that the benefit from noise-handling methods is rather limited and sometimes inconsistent, especially under weak noise. TAPT, on the other hand, is a stable method to assist text classifiers when learning with noisy labels.

\section*{Acknowledgments}
This work has been partially funded by the Deutsche Forschungsgemeinschaft (DFG, German Research Foundation) – Project-ID 232722074 – SFB 1102 and the EU Horizon 2020 projects ROXANNE under grant number 833635 and COMPRISE under grant agreement No. 3081705.

\bibliography{africa_nlp_ref}
\bibliographystyle{iclr2022_conference}

\clearpage
\appendix
\section{Detailed on Model Performance}
\label{appendix:detailed_performance}
We summarize our experimental results on simulated noise and weak supervision noise in Table \ref{tab:performance_noise_simulation} and \ref{tab:performance_weak_noise}, respectively.

\begin{table}[h!]
\centering
\scalebox{0.56}{
\begin{tabular}{@{}llllllllllll@{}}
\toprule
\multirow{3}{*}{} & \multicolumn{7}{c}{AG-News} & \multicolumn{4}{c}{IMDB} \\ \cmidrule(lr){2-8} \cmidrule(lr){9-12} 
 & \multicolumn{1}{c}{} & \multicolumn{3}{c}{uniform} & \multicolumn{3}{c}{single-flip} &  & \multicolumn{3}{c}{single-flip} \\ \cmidrule(lr){3-5} \cmidrule(lr){6-8} \cmidrule(lr){10-12} 
  & \multicolumn{1}{c}{clean} & \multicolumn{1}{c}{40\%} & \multicolumn{1}{c}{60\%} & \multicolumn{1}{c}{70\%} & \multicolumn{1}{c}{20\%} & \multicolumn{1}{c}{40\%} & \multicolumn{1}{c}{45\%} & \multicolumn{1}{c}{clean} & \multicolumn{1}{c}{20\%} & \multicolumn{1}{c}{40\%} & \multicolumn{1}{c}{45\%} \\ \midrule

NV & 94.07$\pm$0.13 & 84.48$\pm$0.78 & 61.61$\pm$3.18 & 43.78$\pm$5.07 & 90.46$\pm$0.37 & 76.06$\pm$0.33 & 64.74$\pm$0.94 & 94.03$\pm$0.13 & 86.34$\pm$0.77 & 65.05$\pm$0.90 & 58.97$\pm$1.26\\
CT & - & 92.18$\pm$0.21 & 89.90$\pm$0.38 & 84.74$\pm$2.56 & 93.33$\pm$0.12 & 90.62$\pm$0.53 & 87.99$\pm$1.64 & - & 92.32$\pm$0.27 & 89.36$\pm$0.67 & 83.77 $\pm$3.88\\
NMat & - & 92.25$\pm$0.14 & 89.91$\pm$0.48 & 83.9$\pm$1.87 & 93.91$\pm$0.15 & 93.13 $\pm$0.31 & 92.93 $\pm$0.51 & - & 92.07$\pm$0.21 & 87.13$\pm$0.44 & 78.82$\pm$1.37 \\
NMwR & 93.64$\pm$0.06 & 92.02$\pm$0.20 & 89.91 $\pm$0.33 & 84.77$\pm$2.24 & 93.03$\pm$0.17 & 90.23$\pm$0.65 & 88.93$\pm$0.68 & 93.68$\pm$0.14 & 92.12$\pm$0.35 & 85.94$\pm$0.86 & 80.17$\pm$2.57 \\
LS & 94.43$\pm$0.19 & 92.45 $\pm$0.21 & 89.79$\pm$0.38 & 86.64$\pm$0.78 & 93.56$\pm$0.23 & 92.40$\pm$0.33 & 90.94$\pm$0.86 & 94.06 $\pm$0.09 & 92.13$\pm$0.43 & 87.22$\pm$1.39 & 80.61$\pm$2.48\\
WN & 94.40$\pm$0.13 & 92.40$\pm$0.25 & 89.53$\pm$0.75 & 85.49$\pm$0.76 & 93.80$\pm$0.08 & 92.33$\pm$0.35 & 88.94$\pm$0.92 & 93.98$\pm$0.15 & 92.13$\pm$0.21 & 85.88$\pm$2.78 & 80.12$\pm$4.09 \\ \midrule
TAPT+NV & 94.71$\pm$0.12 & 82.58$\pm$1.19 & 54.84$\pm$0.96 & 36.56$\pm$0.89 & 90.89$\pm$0.13 & 71.51$\pm$1.00 & 61.02$\pm$1.75 & 94.82$\pm$0.06 & 87.03$\pm$1.55 & 64.01$\pm$1.52 & 58.38$\pm$0.90 \\
TAPT+CT & - & 93.14 $\pm$0.25 & 90.67$\pm$0.27 & 85.64$\pm$2.12 & 94.09$\pm$0.24 & 91.06$\pm$0.51 & 88.16$\pm$2.16 & - & 93.93 $\pm$0.14 & 91.08 $\pm$0.87 & 89.91 $\pm$0.48\\
TAPT+NMat & - & 93.03$\pm$0.31 & 90.79$\pm$0.22 & 86.70$\pm$1.27 & 94.41 $\pm$0.12 & 94.07 $\pm$0.26 & 93.64$\pm$0.14 & - & 92.32$\pm$0.40 & 89.69$\pm$0.47 & 83.93$\pm$2.99 \\
TAPT+NMwR & 94.29$\pm$0.11 & 92.73$\pm$0.19 & 90.43$\pm$0.60 & 87.54$\pm$1.19 & 94.05$\pm$0.08 & 93.50$\pm$0.19 & 93.41$\pm$0.16 & 94.84$\pm$0.08 & 93.58$\pm$0.27 & 89.61$\pm$1.60 & 85.32$\pm$3.05\\
TAPT+LS & 94.81$\pm$0.04 & 92.85$\pm$0.23 & 90.60$\pm$0.53 & 87.23$\pm$0.97 & 94.28$\pm$0.14 & 93.03$\pm$0.53 & 90.80$\pm$1.12 & 94.79$\pm$0.06 & 93.21$\pm$0.45   & 90.27$\pm$1.16 & 80.71$\pm$4.74\\
TAPT+WN & 94.85 $\pm$0.09 & 92.94$\pm$0.18 & 90.96$\pm$0.82 & 87.62 $\pm$0.92 & 94.35$\pm$0.08 & 92.87$\pm$0.61 & 90.94$\pm$1.25 & 94.86 $\pm$0.05 & 93.51$\pm$0.35 & 90.81$\pm$0.69 & 84.66$\pm$3.25 \\ \bottomrule
\end{tabular}}
\caption{Average test accuracy (\%) and standard deviation (5 trials) on AG-News and IMDB with uniform and single-flip noise. NV: without noise-handling and no validation set, i.e. train the model without noise-handling and until the training loss converges. CT: Co-Teaching. NMat: Noise Matrix. NMwR: Noise Matrix with Regularization. LS: Label Smoothing. CT and NMat are equivalent to WN in the clean setting. TAPT+[XX]: Task adaptive pre-training followed by the method XX. Note that as IMDB is a binary-classification task, single-flip noise is equivalent to the uniform noise in this case.}
\label{tab:performance_noise_simulation}
\end{table}

\setlength{\tabcolsep}{3.5pt}
\begin{table}[h!]
\centering \fontsize{6}{6}\selectfont
\begin{tabular}{@{}lllllll@{}}
\toprule
 & \multicolumn{2}{c}{FT} & \multicolumn{2}{c}{TAPT+FT} \\ \cmidrule(lr){2-3} \cmidrule(lr){4-5}
 & \multicolumn{1}{c}{\yoruba} & \multicolumn{1}{c}{Hausa}  & \multicolumn{1}{c}{\yoruba} & \multicolumn{1}{c}{Hausa} &  \\ \midrule
NV & 63.88$\pm$1.59 & 46.98$\pm$1.01 & 66.78$\pm$1.38 & 47.32$\pm$0.85 \\
CT & 61.37$\pm$1.58 & 31.65$\pm$2.71 & 63.35$\pm$1.21 & 36.16$\pm$5.66  \\
NMat & 65.96 $\pm$0.81 & 46.58 $\pm$0.88  & 69.08$\pm$1.56 &  47.37$\pm$0.69 \\
NMwR & 61.32$\pm$0.71 & 35.36$\pm$3.60 & 63.46$\pm$1.44 & 41.07$\pm$3.75 \\
LS & 65.44$\pm$1.67 & 46.36$\pm$0.78  & 69.16$\pm$1.35 & 46.87$\pm$1.06 \\
WN & 64.72$\pm$1.45 & 46.39$\pm$0.81 & 69.58 $\pm$1.58 & 46.44$\pm$0.60 \\ \bottomrule
\end{tabular}
\caption{Average test accuracy (\%) and standard deviation (10 trials) on \yoruba and Hausa with noise from weak supervision. FT: direct fine-tuning on text classification task NV: without noise-handling and no validation set, i.e. train the model without noise-handling and until the training loss converges. CT: Co-Teaching. NMat: Noise Matrix. NMwR: Noise Matrix with Regularization. LS: Label Smoothing. CT and NMat are equivalent to WN in the clean setting.}
\label{tab:performance_weak_noise}
\end{table}

\end{document}